\documentclass[runningheads]{llncs}

\usepackage[final]{eccv}
\usepackage{eccv}

\usepackage{eccvabbrv}
\usepackage[accsupp]{axessibility}  

\usepackage{microtype}
\usepackage{graphicx}
\usepackage{xcolor}
\usepackage{subcaption}
\usepackage{booktabs} 
\usepackage{multirow}
\usepackage{makecell}

\usepackage{hyperref}

\usepackage{orcidlink}
 
\usepackage{amsmath}
\usepackage{amssymb}
\usepackage{mathtools}
\usepackage{algorithm}
\usepackage{algorithmic}

\usepackage{afterpage}

\usepackage{tcolorbox}
\usepackage[frozencache,cachedir=minted-cache]{minted2}
\usepackage{tikz}
\tcbuselibrary{listings,breakable}

\begin{document}

\title{
    Visual Prompt Discovery via Semantic Exploration 
}


\author{Jaechang~Kim\inst{1,2}$^*$\orcidlink{0009-0009-1240-5869} \and
Yotaro~Shimose\inst{1} \and
Zhao~Wang\inst{1}\orcidlink{0000-0002-9169-5391} \and
Kuang-Da~Wang\inst{1,3}$^*$\orcidlink{0009-0004-0846-8254} \and
Jungseul~Ok\inst{2}\orcidlink{0000-0003-4742-2473} \and
Shingo~Takamatsu\inst{1}\orcidlink{0009-0008-1640-2406}
}

\authorrunning{J. Kim et al.}

\institute{Sony Group Corporation \and
Pohang University of Science and Technology \and
National Yang Ming Chiao Tung University \\ 
$^*$ Work done during an internship in Sony \\
\email{jaechang@postech.ac.kr}, \email{Shingo.Takamatsu@sony.com}
}

\maketitle


\begin{abstract}

Large Vision-Language Models (LVLMs) encounter significant challenges in image understanding and visual reasoning, leading to critical perception failures. 
Visual prompts, which incorporate image manipulation code, have shown promising potential in mitigating these issues.
While visual prompts have emerged as a promising direction, previous methods for visual prompt generation have focused on  tool selection rather than diagnosing and mitigating the root causes of LVLM perception failures.
Because of the opacity and unpredictability of LVLMs, optimal visual prompts must be discovered through empirical experiments, which have relied on manual human trial-and-error.

In this work, we propose an automated semantic exploration framework for discovering task-wise visual prompts. 
Unlike previous methods, our approach enables diverse yet efficient exploration through agent-driven experiments, minimizing human intervention and avoiding the inefficiency of per-sample generation. 
We introduce a semantic exploration algorithm named SEVEX, which addresses two major challenges of visual prompt exploration: 
(1) the distraction caused by lengthy, low-level code and 
(2) the vast, unstructured search space of visual prompts. 
Specifically, our method leverages an abstract idea space as a search space, a novelty-guided selection algorithm, and a semantic feedback-driven ideation process to efficiently explore diverse visual prompts based on empirical results.

We evaluate SEVEX on the BlindTest and BLINK benchmarks, which are specifically designed to assess LVLM perception. Experimental results demonstrate that SEVEX significantly outperforms baseline methods in task accuracy, inference efficiency, exploration efficiency, and exploration stability. Notably, our framework discovers sophisticated and counter-intuitive visual strategies that go beyond conventional tool usage, offering a new paradigm for enhancing LVLM perception through automated, task-wise visual prompts.

\end{abstract}

\begin{figure}[t]
    \centering
    \includegraphics[width=\linewidth]{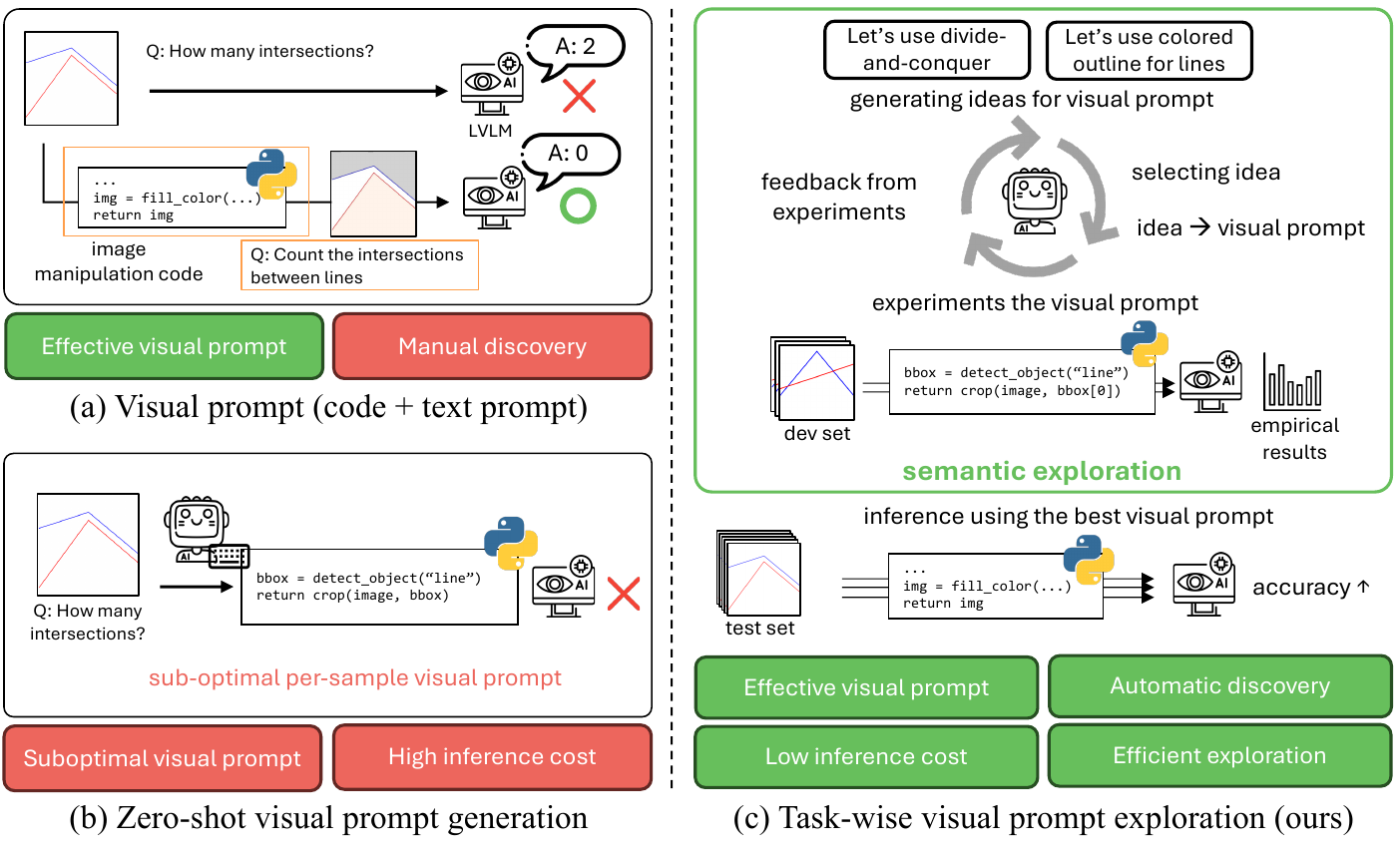}
    \vspace{-5mm}
    \caption{
        \textbf{Overview of visual prompting paradigms for LVLMs.}
        Visual prompts are an effective approach for mitigating perception failure of LVLM.
        Rather than manual discovery and zero-shot generation, we propose a framework to explore task-wise visual prompts efficiently. 
    }
    \label{fig:problem_setting}
\end{figure}

\section{Introduction}
\label{sec:intro}

Large Vision-Language Models (LVLMs) have demonstrated remarkable capabilities in complex reasoning and open-ended dialogue. However, LVLMs often struggle with fundamental visual perception such as identifying fine-grained attributes or understanding spatial relationships, leading to hallucinations or incorrect reasoning based on misperceived visual input~\cite{vlm_blind, vlm_eye_exam, blink}. The inherent opacity and unpredictability of these models make it difficult to anticipate such failures, necessitating specialized interventions to ground their vision more reliably.

One of the most promising directions to mitigate these failures is visual prompting~\cite{visual_prompting_survey, viser, som, VisProg, sketchpad, mllms_know_where, blackbox_vp}. In addition to traditional text-based prompts, visual prompting involves direct manipulation of the input image—often through programmatic code—to explicitly guide the model’s attention to relevant features. By overlaying geometric cues~\cite{viser, blackbox_vp}, highlighting specific regions~\cite{som}, or applying external tools~\cite{sketchpad}, visual prompts enable LVLMs to bypass the failure points and to focus on the most important part. This programmatic approach allows for a precise, reproducible way to enhance visual grounding, offering a powerful tool for correcting the perceptual blind spots of LVLMs.

Despite its potential, current research on visual prompt generation remains in its infancy, often focusing on tool usage rather than diagnosing and resolving fundamental perception failures. More importantly, discovering an effective visual prompt for a specific task remains a heavily manual process of trial-and-error. 
Because LVLMs react to visual changes in highly sensitive and non-intuitive ways, human researchers must spend significant effort empirically testing various ideas to find the optimal strategy.
This challenge is further compounded by the lack of transferability across different architectures; as we demonstrate in Section~\ref{subsec:transferability}, a visual prompt optimized for one backbone model rarely yields consistent gains when applied to another.
This implies that a visual prompt for every model needs to be discovered independently.
This heavy dependency on human effort limits the scalability of visual prompting and underscores the urgent need for an automated, agent-driven discovery framework.

However, automating this discovery process is not straightforward; it presents two significant technical challenges. First, the distraction of low-level code: lengthy and complex image-manipulation scripts can introduce unintended noise, overwhelming the LVLM instead of aiding it. Second, the vast and unstructured search space: the infinite combinations of visual modifications make it nearly impossible for a naive agent to find an optimal solution. 

To address these, we propose an automated semantic exploration framework for task-wise visual prompts. Instead of searching through raw code, our approach operates on a high-level idea space. By utilizing a novelty-guided selection algorithm and backpropagating actionable insights from sample-wise analysis, our framework efficiently explores diverse visual strategies and converges on effective, task-wise prompts with minimal human oversight.

We evaluate our framework on two benchmarks~\cite{blink, vlm_blind} specifically designed to expose the visual perception failures of LVLMs. Our experiments demonstrate that the automatically discovered visual prompts significantly outperform existing baselines. Furthermore, our analysis reveals that optimal visual prompts are rarely transferable across different LVLM settings, a finding that strongly emphasizes the necessity of our automated discovery framework for model-specific optimization. These results validate the efficiency of semantic exploration and suggest a new paradigm for enhancing the reliability of vision-language models through automated discovery.

We illustrate the problem setting and our motivation in Figure~\ref{fig:problem_setting} and summarize our contributions as follows:

\begin{itemize}
    \item \textbf{Automated Discovery of Task-wise Visual Prompts:} We introduce an agent-driven framework that automatically discovers task-wise visual prompts, moving beyond manual engineering and sub-optimal zero-shot generation.
    Recognizing that effective visual prompts are highly model-specific and rarely transferable, our framework provides a scalable solution.
    
    \item \textbf{Semantic Exploration:} We propose SEmantic Visual prompt EXploration (SEVEX) to overcome the challenges of exploration in raw visual prompt space. 
    SEVEX utilizes 1) an abstract idea space as a search space and 2) a novelty-guided node selection algorithm to enable efficient and diverse exploration, and 3) a semantic backpropagation that analyzes sample-wise results into actionable insights for future idea generation, enabling the diverse and efficient exploration.
    
    \item \textbf{Performance \& Analysis:} We demonstrate the effectiveness of SEVEX on the BlindTest and BLINK benchmarks, achieving superior task accuracy, inference efficiency, and exploration stability. Notably, our framework discovers sophisticated, counter-intuitive visual strategies, not limited to conventional tool usage.
    
\end{itemize}

\section{Related Work}
\label{sec:related_work}

\subsection{Perception failures in LVLMs and visual prompting}

\subsubsection{Perception failures in LVLMs.} 
Recent studies have increasingly revealed that LVLMs, despite their advanced reasoning capabilities, suffer from systematic \textit{perception failures}. Benchmarks such as BlindTest~\cite{vlm_blind}, VLM Eye Exam~\cite{vlm_eye_exam}, and Blink~\cite{blink} demonstrate that these models often struggle with fundamental visual tasks, including fine-grained attribute identification and spatial relationship understanding~\cite{VLM_spatial_ambiguity, vlm_spatial_psychometrics}. These failures lead to hallucinations where the model's reasoning is grounded on misperceived visual information. 

\subsubsection{Visual prompting strategies.}
We define a \emph{visual prompt} as image-manipulation code paired with text prompts. Existing strategies for improving LVLM perception can be grouped into two paradigms, depending on their objective and generation timing:
1) \emph{Visual tool-use:} Early methods~\cite{VisProg, sketchpad} generate Python programs at inference time to call external vision tools, such as segmentation or depth estimation models. While effective at outsourcing perception, they treat the LVLM mainly as a controller and do not address its intrinsic visual reasoning failures. Since tool-use is generated zero-shot without diagnosis, an initially misinterpreted problem can lead to mismatched tools, cascading errors, and high inference cost. Fine-tuning-based methods~\cite{wu2025vtoolr1, vtsv} also focus on tool selection rather than understanding when LVLMs fail.
2) \emph{Visual scaffolding for intrinsic perception:} In contrast, visual scaffolding directly modifies the input image to mitigate intrinsic failures such as spatial ambiguity and attribute binding. Methods such as Viser~\cite{viser}, BBVPE~\cite{blackbox_vp}, Zoom-in~\cite{mllms_know_where}, and Set-of-Mark~\cite{som} overlay geometric cues or markers on task-relevant regions. However, these methods remain largely heuristic, manually designed, or dependent on model fine-tuning.

Despite its promise, visual scaffolding remains underexplored as an automated, task- and model-specific prompting strategy. Existing methods are mostly static or require human experimentation, and our experiments show that LVLM sensitivity makes shared visual prompts ineffective across backbones. Our framework fills this gap with an agent-driven system that automatically discovers task-wise visual prompts through empirical experimentation, yielding model-specific prompts that mitigate intrinsic perception failures without per-sample tool generation.

\subsection{Automated prompt engineering}

\subsubsection{From text to visual domain.}
Since the introduction of Chain-of-Thought (CoT) prompting~\cite{step-by-step}, prompt engineering has become essential for maximizing the performance of Large Language Models (LLMs). To overcome the limitations of manual trial-and-error, automated methods~\cite{deepbreath, automatic_prompt_engineer} have been developed to optimize prompts based on empirical feedback from small datasets. In the visual domain, early optimization efforts primarily focused on vision-language representation models like CLIP, utilizing gradient-based soft prompt learning~\cite{clip_soft_visual_prompt, clip_visual_prompt_learning}. However, these continuous optimization techniques are inapplicable to the discrete and non-differentiable nature of generating programmatic visual prompts for LVLMs, which require structured code and natural language instructions rather than continuous vector adjustments.

\subsubsection{Complexity of programmatic visual prompts.}
A significant hurdle in automating visual prompt engineering is the inherent complexity and length of the prompts. Unlike text-only prompts, a programmatic visual prompt consists of a combination of formatted image-manipulation code and descriptive text. This hybrid structure results in significantly longer contexts, which poses a long-context distraction problem for LLMs~\cite{lost_in_the_middle}. The increased search space of potential code snippets and their sensitive interaction with the LVLM's perception make direct generation and optimization via raw text or code search extremely difficult. Direct search often leads to the agent being overwhelmed by low-level implementation details rather than focusing on the high-level logic of visual grounding.

\subsubsection{Mitigating complexity via semantic exploration.}
To mitigate these issues, our framework shifts the paradigm from searching in the raw visual prompt to semantic exploration within an abstract idea space. Instead of forcing the agent to reason over every line of low-level manipulation code, our approach enables the agent to iterate on high-level conceptual strategies. By decoupling the semantic intent (the "Idea") from its implementation (the "Code"), our method reduces the cognitive load on the agent and enables a more efficient, novelty-guided search. This abstraction allows the agent to focus on diagnosing the core perception failures and discovering effective visual strategies, bridging the gap between automated text prompt engineering and the complex requirements of LVLM visual grounding.

\begin{figure*}[t]
    \centering
    \includegraphics[width=\linewidth]{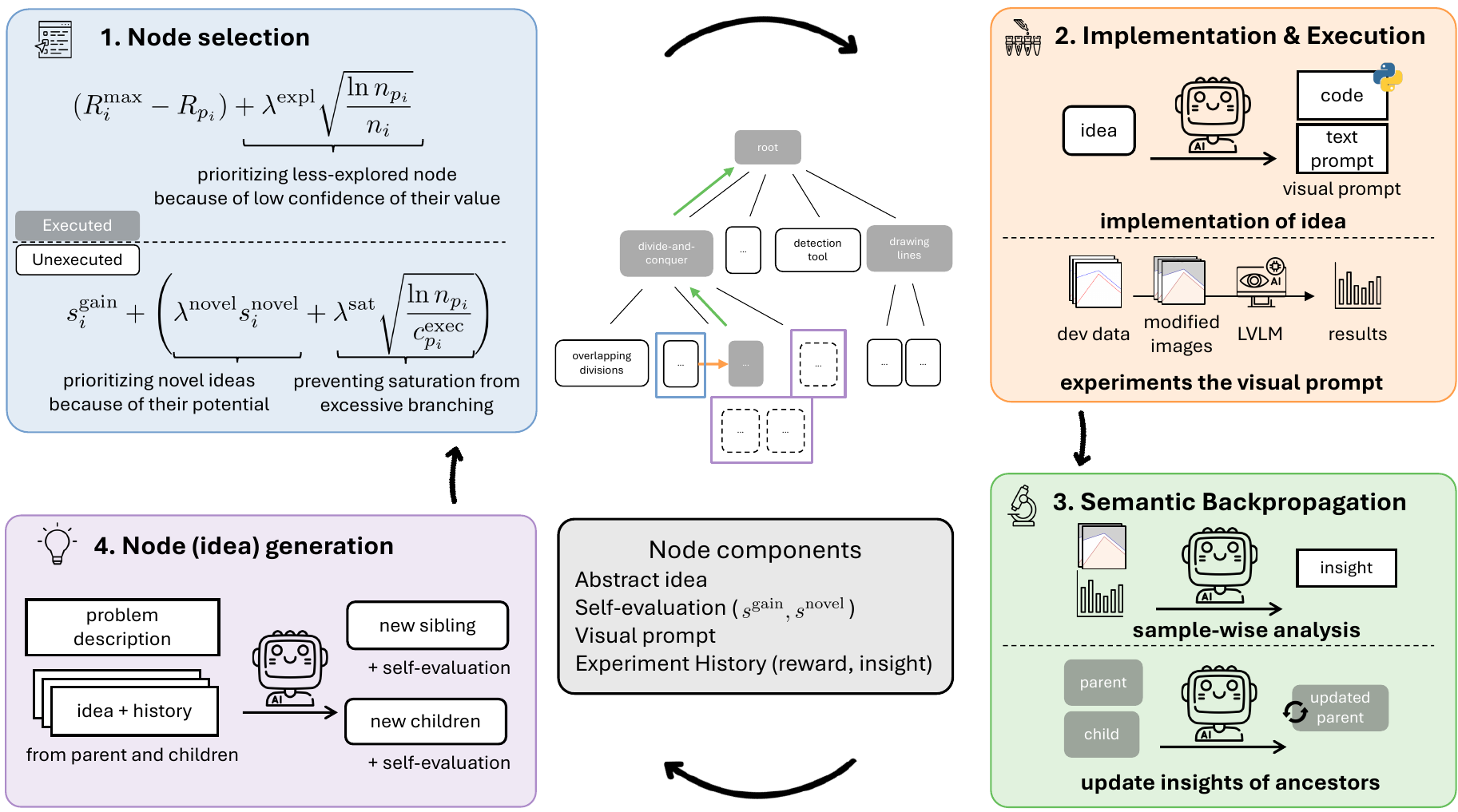}
    \caption{
        \textbf{Overview of SEVEX.} 
        SEVEX generates diverse ideas, selects a node with the highest potential, executes the idea, and back-propagates the insights for future idea generation.
    }
    \label{fig:method}
\end{figure*}

\section{Method}
\label{sec:method}

We propose SEmantic Visual prompt EXploration (SEVEX), which is designed to discover optimal, task-wise visual prompts for LVLMs. Unlike previous works that rely on manual trial-and-error or per-sample tool-use, our framework treats visual prompt discovery as an iterative search problem over a high-level idea space. The core of our approach is a dynamically expanding search tree $\mathcal{T}$ that enables an agent to hypothesize, implement, and explore visual prompts based on empirical feedback. We illustrate the overview of our algorithm in Figure~\ref{fig:method}.

\subsection{Search space and dynamic tree structure}
\label{subsec:design}
To overcome the vast and unstructured nature of visual prompts, we define the search space in a high-level domain. The search tree $\mathcal{T}$ is not pre-defined but grows dynamically as the agent generates new ideas. 
Every executed node has $k$ unexecuted child nodes to ensure exploration of diverse nodes. 
Whenever an unexecuted node is executed, a new node with a new idea is generated.
Each node $N$ in the tree represents a distinct idea for a visual prompt and encapsulates the following components:
\begin{itemize}
    \item \textbf{Abstract Idea ($I$):} A natural language idea describing a visual prompt. 
    \item \textbf{Implementation ($P$):} The concrete instantiation of $I$, consisting of executable Python code utilizing a pre-defined set of visual tools and the corresponding textual prompt.
    \item \textbf{Self-Evaluation Estimates ($S$):} Heuristic scores $\{s^{\text{gain}}, s^{\text{novel}}\}$ evaluated by the agent to guide the search before execution. $s^{\text{gain}}$ represents expected gain from using that idea, and $s^{\text{novel}}$ represents the novelty of the idea compared to its siblings. 
    \item \textbf{Experiment History ($H$):} Results of empirical experiments, including performance metrics and qualitative observations from the experiment and insights back-propagated from descendant nodes.
\end{itemize}

\subsection{Exploration pipeline}
\label{subsec:pipeline}
The exploration follows a four-stage iterative loop: 1) \textbf{Selection}, where the most promising idea is chosen; 2) \textbf{Implementation and Execution}, where the idea is translated into a visual prompt and tested on a development set; and 3) \textbf{Backpropagation}, where results are distilled into insights to guide the next generation of ideas; 4) \textbf{Expansion}, where new ideas are generated based on the insights from the previous executions.

\subsection{Node selection via Novelty-guided UCT}
\label{subsec:selection}

We use the same motivation of Upper Confidence Bound (UCB)~\cite{ucb} algorithm and UCB for Trees (UCT)~\cite{uct} algorithm, which employ node value and its potential for exploration. 
Previous algorithms execute every child node at least once and calculate the potential based on the confidence of node value. 
However, this approach is inefficient in our context because LLM can generate infinite number of new children from a parent node, and generated ideas can be similar to each other or include less promising ideas.

To overcome these challenges, we propose a Novelty-guided UCT (NUCT) which employs novelty to estimate the potential of an unexecuted node. 
NUCT differentiates the priority scoring mechanism based on a node’s execution status, selecting between using empirical results and using expectations. In our tree structure, every executed node maintains $k$ unexecuted child nodes to ensure continuous exploration. 
The novelty of an unexecuted node is evaluated with two terms: a novelty compared to its siblings, and the number of executed siblings, which corresponds to the saturation of the parent node.
Specifically, the priority score $P_i$ for a node $i$ is determined as follows:

\emph{For executed nodes ($n_i > 0$):}
	We follow the standard UCB formula with slight modification. 
    Instead of the mean reward, we use the maximum reward achieved by the node or any of its descendants ($R_{i}^{\max}$) relative to its parent's reward ($R_{p_i}$):

	\begin{equation}
        P_i = (R^{\max}_i - R_{p_i}) + \lambda^\text{expl} \sqrt{\frac{\ln n_{p_i}}{n_i} }
    \end{equation}
    , where $n_i$ is the visitation count and $p_i$ is the parent node of node $i$. 
    This term encourages selecting the node with the best reward in the layer-wise comparison. 

\emph{For unexecuted nodes ($n_i = 0$):}
	Since empirical rewards are unavailable, we estimate the reward and its potential using the agent's self-evaluation and saturation of its parent node:

	\begin{equation}
        P_i = s^\text{gain}_i + \left( \lambda^\text{novel}  s^\text{novel}_i  + \lambda^\text{sat} \sqrt{\frac{\ln n_{p_i}}{c^\text{exec}_{p_i}} }\right)
    \end{equation}
	, where $s^{\text{gain}}_i$ and $s^{\text{novel}}_i$ are normalized scores for the agent's estimate of gain from using the idea and how distinct the idea is from its siblings.
    $c^{exec}_{p_i}$ denotes the number of executed children of $p_i$, which is equivalent to the number of executed sibling nodes of node $i$. 
    This saturation term (multiplied by $\lambda^{\text{sat}}$) penalizes over-explored branches, prioritizing deeper exploration of promising descendants rather than executing less-promising child ideas of a well-explored parent.

\subsection{Implementation and empirical evaluation}
Once a node is selected, the agent acts as an ``Engineer'' to instantiate the abstract idea ($I$) into a visual prompt ($P$). We provide the agent with a list of visual tools and documentation to use them. This visual prompt is executed on a development set, which is a small subset for exploration. The performance (e.g., accuracy) and intermediate images are saved, and numerical score is saved as the empirical reward $R_i$.

\subsection{Semantic backpropagation with sample-wise analysis}
\label{subsec:backprop}
Following execution, an Analyst agent performs a sample-wise failure analysis. Rather than simply propagating numerical rewards, we implement \textbf{Semantic Backpropagation}. 
The analyst diagnoses why a strategy succeeded or failed with executed results for development set. 
We provide predictions and ground truth for each sample, and a few representative images actually passed to the LVLM. To reduce token consumption during exploration, we use representative images for successful cases and failed cases instead of using all images.
These raw analyses are distilled into \textbf{Actionable Insights}—high-level lessons about which visual components are effective for the task. These insights are back-propagated to the \textit{Experiment History ($H$)} of all ancestor nodes. This ensures that the important lessons are propagated, preventing the agent from repeating ineffective visual manipulations in future branches.

\subsection{Insight-driven idea generation}
The cycle concludes with the expansion of $\mathcal{T}$. Guided by the updated $H$, the agent generates new child nodes. To ensure a balance between depth and breadth, for every executed node, the agent generates: sibling nodes to explore alternative conceptual directions at the same abstraction level, and child nodes to refine and specialize the current strategy based on backpropagated feedback. 
Newly created nodes are initialized with self-evaluation scores $S$, making them ready for the next Selection phase. 
Agent generates self-evaluation of expectation, novelty, and feasibility. Feasibility evaluates if the idea is feasible to implement with the given tools. If not, the idea is rejected and a new node is generated instead.
Expectation scores and novelty scores are normalized and re-scaled.
This closed-loop system allows the framework to autonomously navigate the complex landscape of visual prompting without human intervention.


\section{Experiments}

\begin{table}[t]
\caption{
    \textbf{Description of Visual tools used in our experiments.}
}
\label{tab:visual_tools}
\centering
\resizebox{\textwidth}{!}{%
\begin{tabular}{l l l l}
\toprule
\textbf{Category} & \textbf{Tool Name} & \textbf{Description} & \textbf{Parameter(s)} \\
\midrule
\multirow{4}{1em}{basic}
  & get\_image\_size             & Returns image dimensions                   & image \\
  & convert\_image\_grayscale    & Converts image to grayscale                & image \\
  & crop                         & Crops a specified region of the image      & image, coordinates \\
  & overlay\_images              & Overlays two images                        & image1, image2, coordinates, opacity \\
\midrule
\multirow{3}{1em}{drawing}
  & draw\_line                   & Draws a line on the image                  & image, coordinates, color \\
  & draw\_box                    & Draws a rectangle on the image             & image, coordinates, color \\
  & draw\_filled\_box            & Draws a filled rectangle on the image      & image, coordinates, color \\
\midrule
\multirow{4}{1em}{external model}
  & detect\_objects              & Object detection using GroundingDINO~\cite{grounding_dino}         & image, text query, threshold \\
  & sliding\_window\_detection   & Sliding window object detection            & image, text query \\
  & segment\_and\_mark           & Semantic segmentation using SemanticSAM~\cite{semantic_sam}    & image, granularity, mark type \\
  & estimate\_depth              & Depth estimation using DepthAnything~\cite{depthanything}       & image \\
\midrule
LVLM
  & ask\_to\_LVLM                 & Sends a query to LVLM and returns response & images, text prompt \\
\bottomrule
\end{tabular}
}
\end{table}

\begin{table*}[t]
\caption{
    \textbf{Comparison of visual prompting methods.} 
    \textbf{Bold} and \underline{underlined} texts denote the best and second-best methods for each setting, respectively.
    Inf. cost indicates the number of total tokens per-sample.
    }
\label{tab:main_table}
\centering

\vspace{-1mm}
\resizebox{0.75\textwidth}{!}{%

\begin{tabular}{l cc cc cc}
\toprule
\multirow{2}{*}{\textbf{Task}} & \multicolumn{2}{c}{\textbf{Naive}} & \multicolumn{2}{c}{\textbf{SketchPad}} & \multicolumn{2}{c}{\textbf{SEVEX(ours)}} \\
\cmidrule(lr){2-3} \cmidrule(lr){4-5} \cmidrule(lr){6-7} 
 & \textbf{accuracy} & \textbf{inf. cost}  & \textbf{accuracy} & \textbf{inf. cost} & \textbf{accuracy} & \textbf{inf. cost}  \\
\midrule
LineIntersections  & \underline{73.0} & 981   & 33.3 & 16730  & \textbf{90.5} & 991   \\
CircledLetter      & 78.8 & 1323  & \underline{82.0} & 13902  & \textbf{83.7} & 1337  \\
SubwayMap          & \underline{60.0} & 1333  & 58.1 & 16482  & \textbf{62.8} & 1585  \\
OverlappingShapes  & \underline{50.4} & 2429  & 16.3 & 24833  & \textbf{52.4} & 2612  \\
\midrule
Average-BlindTest  & \underline{65.6} & 1517 & 47.4 & 13987  & \textbf{72.4} & 1631  \\
\midrule
Jigsaw             & \underline{75.8} & 869   & 70.8 & 14896  & \textbf{95.8} & 682   \\
Depth              & 83.0 & 1349  & \textbf{85.1} & 14158  & \textbf{85.1} & 1457  \\
Spatial            & 81.4 & 1525  & \underline{85.8} & 15818 & \textbf{86.7} & 1555  \\
SemanticCorr.      & 55.0 & 712   & \underline{58.7} & 9477  & \textbf{63.3} & 1358  \\
VisualCorr.        & 87.3 & 641   & \textbf{90.9} & 14291  & \underline{89.4} & 801   \\
\midrule
Average-Blink      & 76.5 & 1019 & \underline{78.3} & 13728  & \textbf{84.1} & 1171  \\
\midrule 
\textbf{Average}   & \underline{71.6} & 1240 & 64.6 & 15621  & \textbf{78.9} & 1375  \\
\bottomrule
\end{tabular}

}

\vspace{1mm}
\caption{
    \textbf{Comparison to visual prompt exploration baseline.} 
    Expl. cost indicates the number of total tokens per-iteration in exploration stage.
    Inf. cost indicates the number of total tokens per-sample in inference stage.
    }
\label{tab:main_table_exploration}
\vspace{-1mm}
\centering
\resizebox{0.9\textwidth}{!}{%

\begin{tabular}{l cccc cccc}
\toprule
\multirow{2}{*}{\textbf{Task}} & \multicolumn{4}{c}{\textbf{SketchPad$+$APE}} & \multicolumn{4}{c}{\textbf{SEVEX(ours)}} \\
\cmidrule(lr){2-5} \cmidrule(lr){6-9} 
 &  \textbf{dev acc.} & \textbf{test acc.} &  \textbf{\makecell{expl. cost }} & \textbf{\makecell{inf. cost }} & \textbf{dev acc.} & \textbf{test acc.} &  \textbf{\makecell{expl. cost }} & \textbf{\makecell{inf. cost }}   \\
\midrule
LineIntersections    & 30.0 & 16.4 & 797$k$ & 11005 & 93.3 & 90.5 & 39$k$ & 991   \\
CircledLetter        & 93.3 & 73.7 & 980$k$ & 17647 & 85.6 & 83.7 & 79$k$ & 1337  \\
SubwayMap            & 50.5 & 42.5 & 759$k$ & 21471 & 80.0 & 62.8 & 165$k$ & 1585  \\
OverlappingShapes    & 33.3 & 22.9 & 818$k$ & 20672 & 56.7 & 52.4 & 86$k$ & 2612  \\
\midrule
Average-BlindTest    & 51.7 & 38.9 & 838$k$ & 17699 & 78.9 & 72.4 & 92$k$ & 1631  \\
\midrule
Jigsaw               & 76.7 & 66.4 & 761$k$ & 18176 & 96.7 & 95.8 & 75$k$ & 682   \\
Depth                & 76.7 & 85.9 & 794$k$ & 18867 & 85.6 & 85.1 & 87$k$ & 1457  \\
Spatial              & 90.0 & 76.2 & 589$k$ & 19074 & 91.1 & 86.7 & 83$k$ & 1555  \\
SemanticCorr.        & 76.7 & 52.6 & 385$k$ & 12354 & 72.2 & 63.3 & 74$k$ & 1358  \\
VisualCorr.          & 83.3 & 83.1 & 761$k$ & 11937 & 88.9 & 89.4 & 78$k$ & 801   \\
\midrule
Average-Blink        & 80.7 & 72.8 & 658$k$ & 16082 & 86.9 & 84.1 & 80$k$ & 1171  \\
\midrule
\textbf{Average}     & 67.8 & 57.7 & 738$k$ & 16800 & 83.3 & 78.9 & 85$k$ & 1375  \\
\bottomrule
\end{tabular}

}
\end{table*}

\begin{figure*}[t]
    \centering
    \begin{minipage}{\linewidth}
        \centering
        \includegraphics[width=0.85\linewidth]{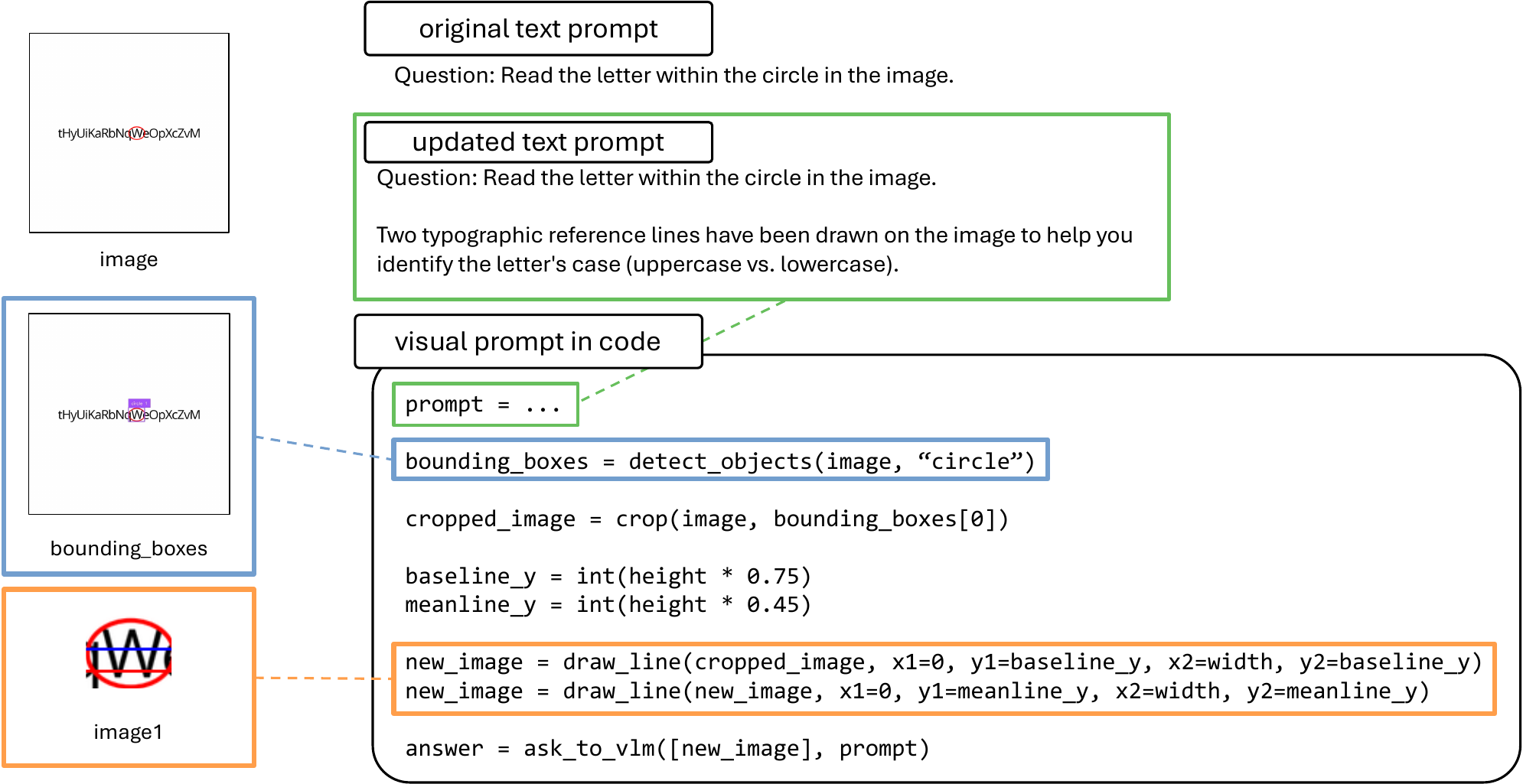}
        \subcaption{
            \textbf{An example visual prompt in CircledLetter.}
            It crops the target region and draws lines for distinguishing uppercase and lowercase.
        }
    \end{minipage}
    \vspace{2mm}
    \begin{minipage}{\linewidth}
        \centering
        \includegraphics[width=0.85\linewidth]{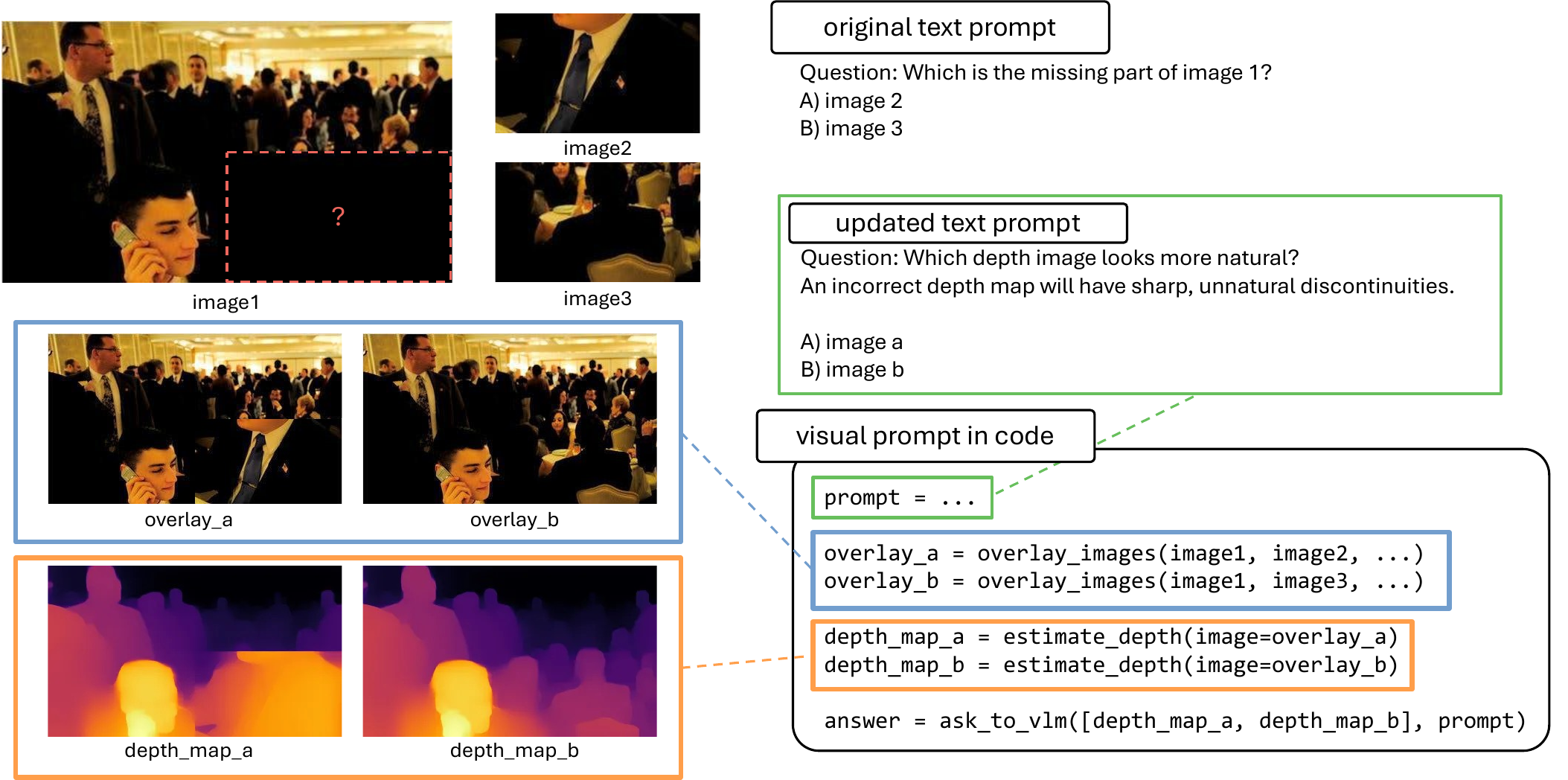}
        \subcaption{
            \textbf{An example visual prompt in Jigsaw.} 
            It overlays images and uses depth estimation model to judge naturalness of images, although it is not the intended usage of the tool.
        }
    \end{minipage}
    \caption{
        \textbf{Results of SEVEX.}
        The code and text prompts are simplified for better visualization.
    } 
    \label{fig:qualitative_prompt}
\end{figure*}

\subsection{Experimental settings}

\subsubsection{Datasets.}
We evaluate our framework using the \textbf{BlindTest}~\cite{vlm_blind} and \textbf{BLINK}~\cite{blink} datasets, which are specifically designed to assess the perception failures of LVLMs. Following the experimental protocol of SketchPad~\cite{sketchpad}, we select a set of tasks requiring fine-grained visual grounding and complex reasoning: 
\begin{itemize} 
    \item \textbf{BlindTest:} Counting Line Intersection, The Circled Letter, Following Single-colored Paths (Subway Map), Counting Overlapping Shapes. 
    \item \textbf{BLINK:} Jigsaw, Relative Depth, Spatial reasoning, Semantic Correlation, Visual Correlation. 
\end{itemize} 
For each task, we randomly sampled 30 images to construct a development set for exploration, reserving the remaining images for the test set. 
Utilizing a small-sized development set aims to demonstrate that SEVEX's semantic exploration is significantly more cost-efficient than alternative optimization methods such as fine-tuning.
We describe the details of each task and the full list of development set samples in Appendix~\ref{appendix:dataset_details}.

\subsubsection{Baseline methods. }
We compare SEVEX against the following visual prompting baselines:
\begin{itemize}
    \item \textbf{Naive}: The standard LVLM inference where the LVLM receives the raw image and the task query without additional visual prompting.
    
    \item \textbf{SketchPad}~\cite{sketchpad}: A zero-shot test-time reasoning framework that dynamically generates image-annotation code during inference. To ensure a fair comparison, we add drawing tools to the original tool list to match the capabilities of SEVEX. SketchPad includes few-shot examples of tool usage and Blink tasks.
    
    \item \textbf{SketchPad$+$APE}~\cite{automatic_prompt_engineer}: An enhanced version of SketchPad where the text prompt is optimized using Automatic Prompt Engineering (APE). We applied iterative APE with resampling, using the original SketchPad prompt as the initial state. This baseline uses the same development set as SEVEX for a controlled comparison.
\end{itemize}

\subsubsection{Implementation details.}
We utilize a comprehensive set of visual tools, which are listed in Table~\ref{tab:visual_tools}. 
We use {Gemini-2.5-flash} as our primary backbone LVLM. We disabled the `thinking' feature during the benchmark inference, while enabled it for SEVEX (e.g., idea generation and implementation) to leverage the agent's reasoning capabilities. According to the documentation of Gemini, disabling thinking yields better results in image perception tasks like image segmentation.
Semantic exploration was conducted for 50 iterations with the following hyperparameters: $\lambda^\text{expl}=0.5$, $\lambda^\text{novel}=0.15$, $\lambda^\text{sat}=0.5$, and $k=3$.

\subsubsection{Evaluation metrics.}
The primary metrics are task accuracy and inference cost, where the cost is defined as the sum of input, output, and reasoning tokens. Unless otherwise specified, all reported accuracies are based on the test set.

\subsection{Comparison to visual prompting methods}

As shown in Table~\ref{tab:main_table} and Table~\ref{tab:main_table_exploration}, 
SEVEX demonstrates superior performance across all critical dimensions: task accuracy, inference efficiency, exploration efficiency, and exploration stability compared to SketchPad and SketchPad$+$APE.

\subsubsection{Task performance.} SEVEX outperforms existing baselines in seven out of nine tasks of BLINK and BlindTest. SEVEX achieves a superior average accuracy of {78.9\%}, significantly surpassing the Naive (71.6\%) and SketchPad (64.6\%) approaches. This performance gap is most pronounced in the BlindTest benchmark, where SEVEX attains an average accuracy of 72.4\% compared to SketchPad's 47.4\%.
The failure of SketchPad in BlindTest can be attributed to its reliance on zero-shot tool use without an empirical understanding of the LVLM's specific perceptual behaviors.
When an LVLM fails at a seemingly trivial task such as counting line intersections, zero-shot generation methods typically fail to detect or recover from such errors, whereas SEVEX identifies effective strategies through systematic experimentation.

\subsubsection{Inference and exploration efficiency.} SEVEX demonstrates remarkable inference efficiency by amortizing the cost of prompt generation over the exploration stage. Consequently, its inference cost is only 10.9\% higher than the Naive method, while providing a 91.2\% reduction in token consumption compared to SketchPad. Furthermore, SEVEX exhibits superior exploration efficiency; its average exploration cost is 11.5\% of the cost required by SketchPad+APE. SEVEX incurs an upfront exploration cost equivalent to approximately 3,442 naive inferences or 273 SketchPad inferences. Consequently, for tasks requiring more than 273 inferences, SEVEX is a more cost-effective solution than SketchPad, while providing the highest accuracy. 
We provide a more detailed comparison to other exploration strategies in Section~\ref{subsec:ablation}.

\subsubsection{Stability of semantic exploration.} 
Despite utilizing a small-sized development set, SEVEX exhibits a narrower generalization gap than SketchPad$+$APE. 
The significant generalization gap observed in SketchPad$+$APE is expected, as it tends to make subtle differences such as paraphrasing.
In contrast, SEVEX maintains higher stability by anchoring its search in high-level ideas first, rather than focusing on the subtle implementation details.
To further examine sensitivity to the choice of development set, we repeat the discovery on additional random development splits for CircledLetter; the discovered strategy remains stable on the held-out test set across splits (see Appendix~\ref{appendix:dev_split}).


\begin{figure}[t]
    \centering    
    \includegraphics[width=0.8\linewidth]{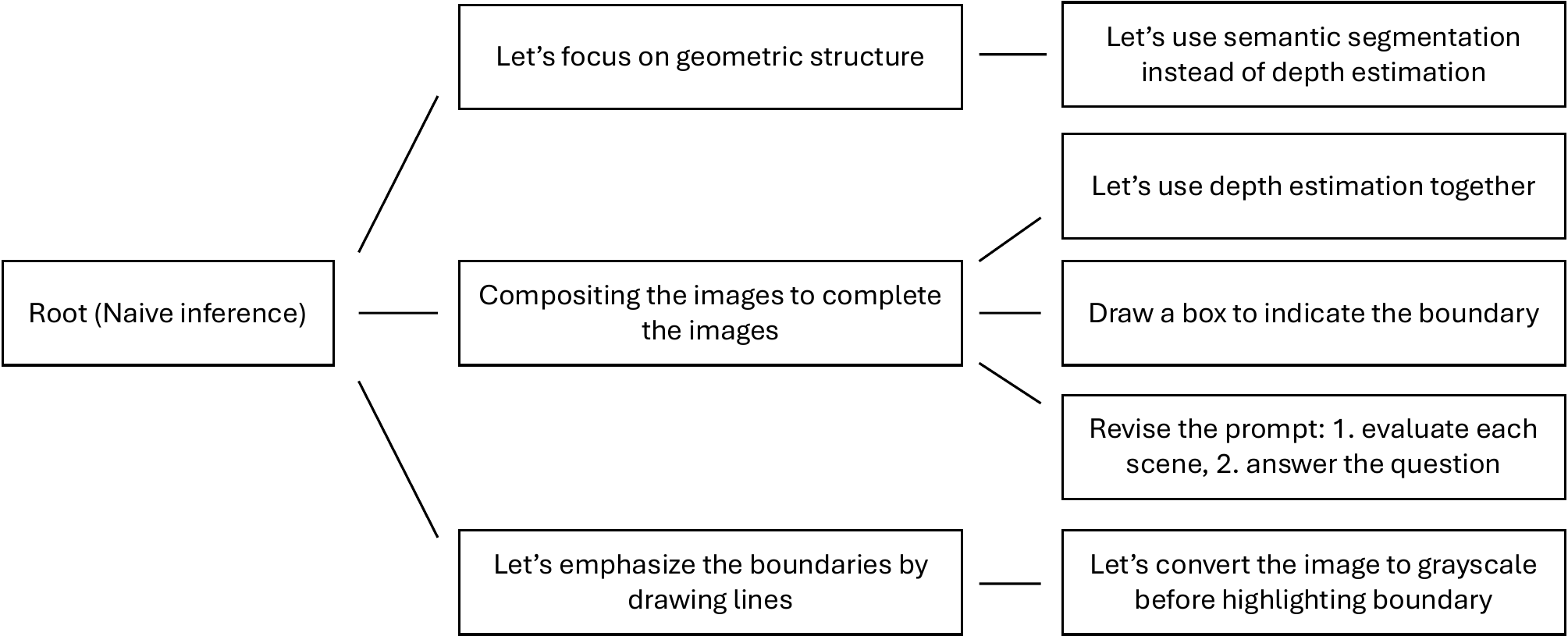}
    \caption{
        \textbf{An example of ideas in a search tree.}
        It shows which kinds of ideas are explored for Jigsaw task.
        A small portion of nodes is visualized for simplicity.
    }
    \label{fig:tree_qualitative}
\end{figure}

\begin{figure}[t]
    \centering
    \begin{minipage}{0.49\linewidth}
        \includegraphics[width=\linewidth]{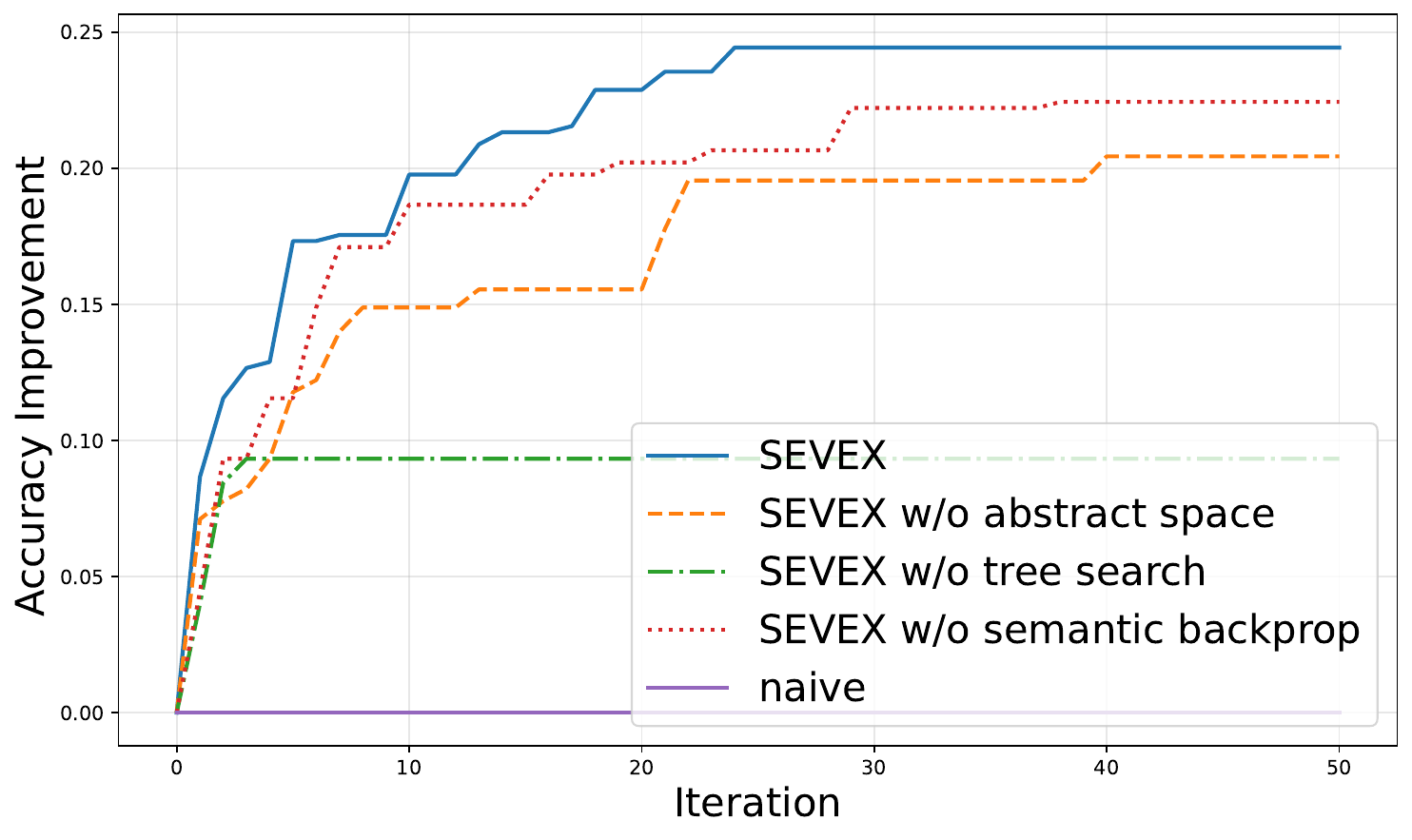}
        \subcaption{Jigsaw}
    \end{minipage}
    \hfill
    \begin{minipage}{0.49\linewidth}
        \includegraphics[width=\linewidth]{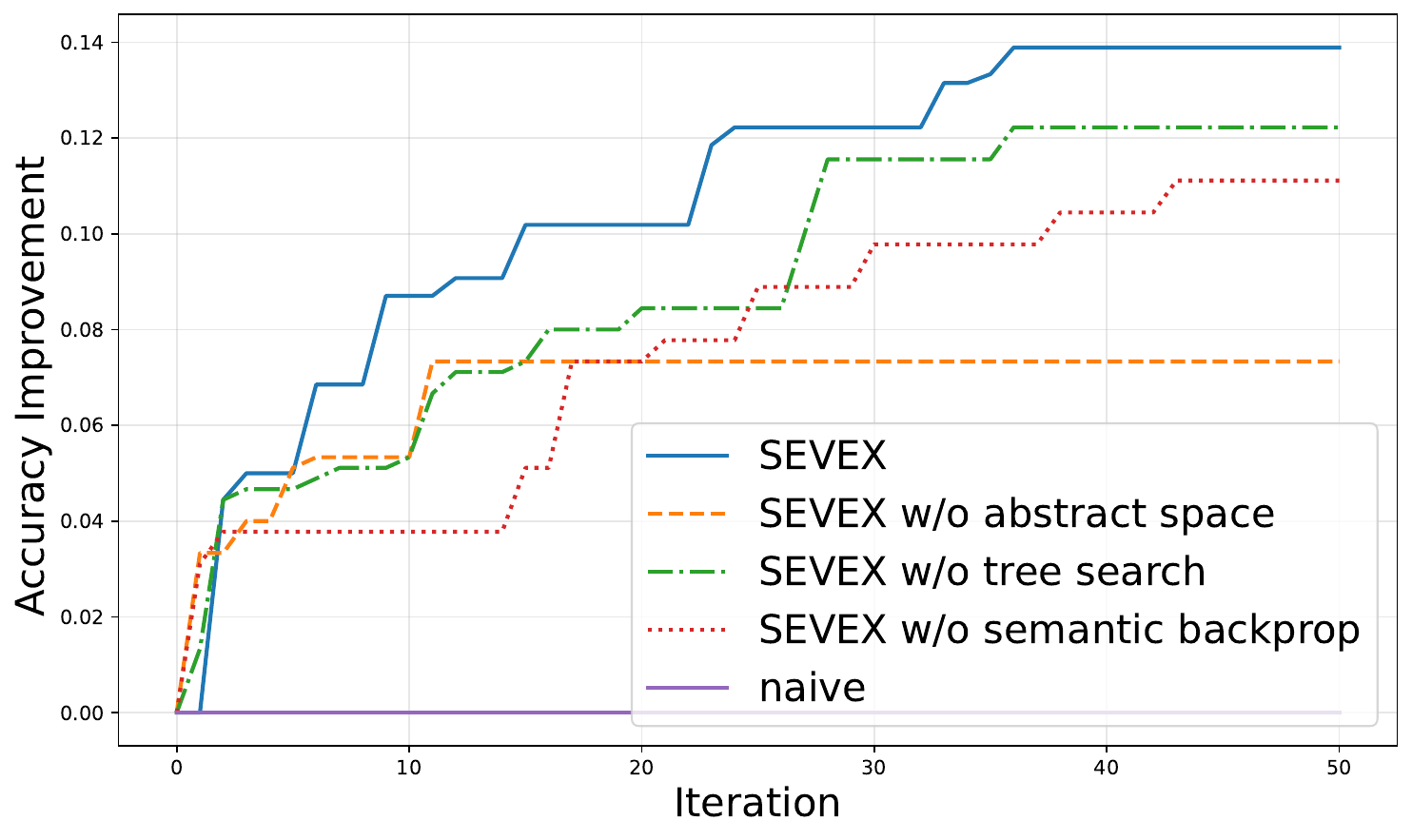}
        \subcaption{CircledLetter}
    \end{minipage}
    \caption{
        \textbf{Accuracy improvements over iterations.} 
        Each figure compares exploration algorithms over iterations.
        We report the accuracies in the development set averaged on five independent runs.
    }
    \label{fig:ablation_efficiency}
\end{figure}


\begin{table}[t]
\caption{
    \textbf{Non-transferability of visual prompts}.
    We compare four visual prompts with three LVLM models.
    Accuracies are calculated in LineIntersection task.
}
\label{tab:transfer}
\centering
\resizebox{0.85\linewidth}{!}{
\begin{tabular}{lccc}
\toprule

 & \makecell{accuracy of \\ Gemini-2.5-flash} & \makecell{accuracy of \\ Claude-Sonnet-4} & \makecell{accuracy of \\ GPT-4o }  \\ 
 \midrule

naive & 73.0 & 66.3 & 8.2 \\
visual prompt 1 (found with Gemini) & 90.5 & 82.9 & 59.1 \\
visual prompt 2 (found with Claude) & 87.9 & 87.8 & 31.0 \\
visual prompt 3 (found with GPT) & 62.3 & 35.0 & 57.1 \\

\bottomrule
\end{tabular}%
}
\vspace{1mm}

    \centering
    \includegraphics[width=.85\linewidth]
    {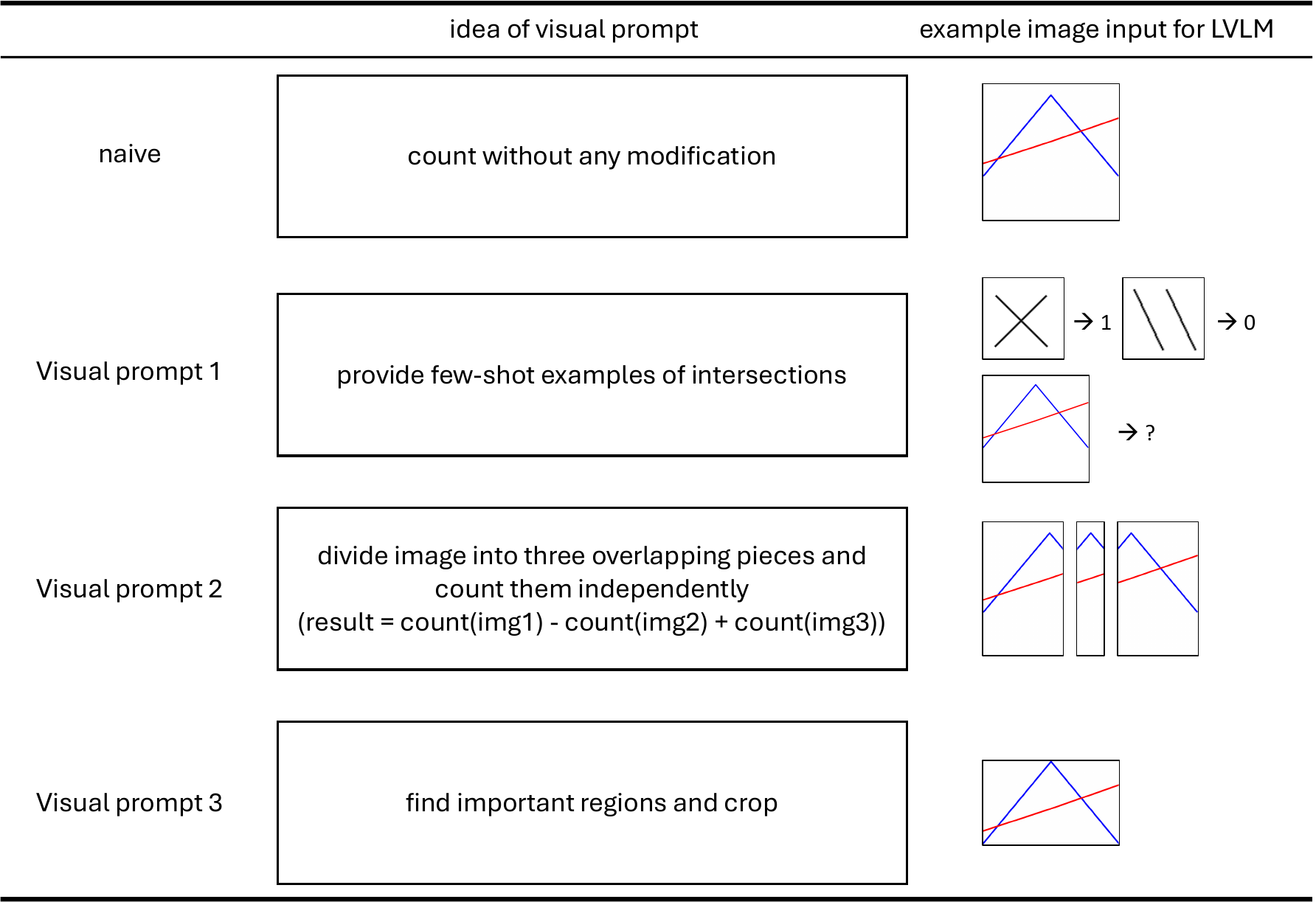}
    \vspace{3mm}
    \captionof{figure}{
        \textbf{Simplified visual prompts used in Table~\ref{tab:transfer}.}
        Each visual prompt is result of SEVEX with different LVLM models.
        Visual prompts have different effects on different LVLM model, which means we cannot expect transferability of visual prompts.
    }
    \label{fig:qualitative_transfer}
    \vspace{-3mm}

\end{table} 

\subsection{Qualitative analysis of visual prompt and exploration tree}

We provide qualitative examples of visual prompts discovered by SEVEX in Figure~\ref{fig:qualitative_prompt}. 
In the \textit{CircledLetter} task, the agent utilizes object detection to localize the target region and renders precise typographic reference lines to assist the LVLM in distinguishing character cases, demonstrating the framework's ability to optimize precise parameters for visual tools.
In the \textit{Jigsaw} task, the agent discovers a counter-intuitive strategy: it overlays missing image components onto the query image and employs a depth estimation model to detect unnatural discontinuities. By repurposing tools in ways not originally intended, the discovery process proves that it is not limited by human bias.
Figure~\ref{fig:tree_qualitative} visualizes a representative exploration tree with its ideas. 
SEVEX first explores diverse conceptual directions at the root level before refining these concepts into concrete implementations through a hierarchical structure. This approach enables the framework to both explore a broad strategy space and refine implementation details based on empirical feedback.

\subsection{Ablation study over exploration strategies}
\label{subsec:ablation}

The contribution of individual exploration components is analyzed in Figure~\ref{fig:ablation_efficiency}. 
We compare SEVEX against: (1) \textit{Naive}, (2) \textit{SEVEX w/o Tree Structure}, and (3) \textit{SEVEX w/o Semantic Backpropagation}, (4) \textit{SEVEX w/o Abstract Space}. 
The results indicate each component is essential for efficiency of SEVEX.
Especially for SEVEX w/o tree structure, it could achieve good performance when the first direction is promising, but if the task is complicated and requires trials in diverse directions, it is less likely to find a good solution.
Because of the resource consumption of each experiment, we only provide ablation studies for Jigsaw and CircledLetter tasks.
To further isolate the contribution of the search itself from that of the external pretrained tools, we additionally evaluate SEVEX with a restricted toolbox containing only basic and drawing primitives; even without external model tools, SEVEX retains most of its gains (see Appendix~\ref{appendix:restricted_toolbox}).

\begin{table}[t]
\caption{
    \textbf{Generality across LVLM backbones.}
    Test accuracy (\%) on four BlindTest tasks for three backbones.
    The Gemini-2.5-flash columns are reproduced from Table~\ref{tab:main_table}; the GPT-4o and Claude Sonnet 4.0 columns are obtained by running the full SEVEX discovery on each backbone.
    The best result within each backbone is in \textbf{bold}.
}
\label{tab:backbone}
\centering
\resizebox{\linewidth}{!}{%
\begin{tabular}{lccccccccc}
\toprule
 & \multicolumn{3}{c}{\textbf{Gemini-2.5-flash}} & \multicolumn{3}{c}{\textbf{GPT-4o}} & \multicolumn{3}{c}{\textbf{Claude Sonnet 4.0}} \\
\cmidrule(lr){2-4}\cmidrule(lr){5-7}\cmidrule(lr){8-10}
\textbf{Task} & Naive & SketchPad & SEVEX & Naive & SketchPad & SEVEX & Naive & SketchPad & SEVEX \\
\midrule
LineIntersections & 73.0 & 33.3 & \textbf{90.5} & 8.2 & 19.2 & \textbf{57.1} & 66.3 & 63.6 & \textbf{87.8} \\
CircledLetter & 78.8 & 82.0 & \textbf{83.7} & 61.1 & 43.9 & \textbf{85.4} & 76.1 & \textbf{90.7} & 80.5 \\
SubwayMap & 60.0 & 58.1 & \textbf{62.8} & 49.6 & 19.6 & \textbf{61.0} & 59.7 & 37.4 & \textbf{60.6} \\
OverlappingShapes & 50.4 & 16.3 & \textbf{52.4} & 36.4 & 33.3 & \textbf{46.2} & 65.1 & 59.8 & \textbf{72.0} \\
\midrule
\textbf{Average} & 65.6 & 47.4 & \textbf{72.4} & 38.8 & 29.0 & \textbf{62.4} & 66.8 & 62.9 & \textbf{75.2} \\
\bottomrule
\end{tabular}%
}
\end{table}

\subsection{Generality across LVLM backbones}
\label{subsec:backbone}

To verify that the gains of SEVEX are not specific to Gemini-2.5-flash, we run the full discovery process on two additional backbones, GPT-4o and Claude Sonnet 4.0, over four BlindTest tasks (Table~\ref{tab:backbone}).
SEVEX improves the average accuracy over the Naive baseline on both GPT-4o (38.8\% $\rightarrow$ 62.4\%) and Claude Sonnet 4.0 (66.8\% $\rightarrow$ 75.2\%), confirming that the improvement persists across backbones with distinct perceptual behaviors.
The gain is largest where the backbone is weakest, such as GPT-4o on LineIntersections (8.2\% $\rightarrow$ 57.1\%).
SEVEX is not uniformly the best on every task and backbone, e.g., SketchPad slightly outperforms SEVEX on Claude for CircledLetter (90.7\% vs. 80.5\%), but SEVEX consistently provides the strongest average.
Crucially, the most effective prompt differs across backbones, which motivates the model-specific discovery analyzed next.

\subsection{Non-transferability of visual prompt}
\label{subsec:transferability}

Table~\ref{tab:transfer} and Figure~\ref{fig:qualitative_transfer} demonstrate that the discovered visual prompts are rarely transferable across different LVLM backbones.
For instance, the \emph{visual prompt 2} that divides an image into three overlapping pieces  yields greater improvements than \emph{visual prompt 3} for Claude, whereas this tendency is reversed for GPT. These results underscore the necessity of discovering model-specific visual prompts tailored to the unique perceptual biases of each backbone. Given the laborious nature of manual prompt engineering, this highlights the importance of our automated discovery framework.

\section{Conclusion}

In this work, we introduced SEVEX, an agent-driven framework designed to automatically discover task-wise visual prompts that mitigate the intrinsic perception failures of  LVLMs. 
Motivated by the observation that LVLMs often struggle with fundamental visual reasoning despite their advanced linguistic capabilities, our approach shifts from manual trial-and-error or per-sample code generation to a structured search within a high-level, abstract idea space. By decoupling semantic intent from programmatic implementation, SEVEX effectively navigates the vast search space of visual modifications while avoiding long-context distraction.

Our experiments on the BlindTest and BLINK benchmarks demonstrate that SEVEX significantly outperforms existing baselines, including zero-shot tool-use frameworks and automated text prompt engineering. A critical takeaway from our analysis is the non-transferability of visual prompts; optimal visual cues discovered for one model frequently fail or even degrade performance when applied to another. This finding reinforces the necessity of an automated, model-specific discovery process to account for the unique perceptual biases of different backbones. 
Finally, we discuss broader implications and future research directions in Appendix~\ref{sec:discussion}.

\section*{Acknowledgement}

Jaechang Kim and Jungseul Ok were partly supported by Institute of Information \& communications Technology Planning \& Evaluation(IITP) grant funded by the Korea government(MSIT)(RS-2024-00509258, Global AI Frontier Lab).

\bibliographystyle{splncs04}
\bibliography{reference}

@article{vlm_eye_exam,
  title={VLM’s Eye Examination: Instruct and Inspect Visual Competency of Vision Language Models},
  author={Nam, Hyeon-Woo and Moon, Ye-Bin and Choi, Wonseok and Lee, Hyun and Oh, Tae-Hyun},
  journal={Transactions on Machine Learning Research},
  volume={2025},
  year={2025},
  publisher={Journal of Machine Learning Research (JMLR)}
}

@inproceedings{viser,
      title={Visual Structures Help Visual Reasoning: Addressing the Binding Problem in VLMs}, 
      author={Amirmohammad Izadi and Mohammad Ali Banayeeanzade and Fatemeh Askari and Ali Rahimiakbar and Mohammad Mahdi Vahedi and Hosein Hasani and Mahdieh Soleymani Baghshah},
      year={2025},
    booktitle = {NeurIPS},
}

@article{sketchpad,
  title={Visual sketchpad: Sketching as a visual chain of thought for multimodal language models},
  author={Hu, Yushi and Shi, Weijia and Fu, Xingyu and Roth, Dan and Ostendorf, Mari and Zettlemoyer, Luke and Smith, Noah A and Krishna, Ranjay},
  journal={Advances in Neural Information Processing Systems},
  volume={37},
  pages={139348--139379},
  year={2024}
}

@article{som,
      title={Set-of-Mark Prompting Unleashes Extraordinary Visual Grounding in GPT-4V}, 
      author={Jianwei Yang and Hao Zhang and Feng Li and Xueyan Zou and Chunyuan Li and Jianfeng Gao},
      journal={arXiv preprint arXiv:2310.11441},
      year={2023},
}

@inproceedings{blink,
  title={Blink: Multimodal large language models can see but not perceive},
  author={Fu, Xingyu and Hu, Yushi and Li, Bangzheng and Feng, Yu and Wang, Haoyu and Lin, Xudong and Roth, Dan and Smith, Noah A and Ma, Wei-Chiu and Krishna, Ranjay},
  booktitle={European Conference on Computer Vision},
  pages={148--166},
  year={2024},
  organization={Springer}
}

@inproceedings{
  mllms_know_where,
  title={{MLLM}s Know Where to Look: Training-free Perception of Small Visual Details with Multimodal {LLM}s},
  author={Jiarui Zhang and Mahyar Khayatkhoei and Prateek Chhikara and Filip Ilievski},
  booktitle={The Thirteenth International Conference on Learning Representations},
  year={2025},
  url={https://arxiv.org/abs/2502.17422}
}

@inproceedings{vlm_blind,
  title={Vision language models are blind},
  author={Rahmanzadehgervi, Pooyan and Bolton, Logan and Taesiri, Mohammad Reza and Nguyen, Anh Totti},
  booktitle={Proceedings of the Asian Conference on Computer Vision},
  pages={18--34},
  year={2024}
}

@article{visual_prompting_survey,
  title={Visual prompting in multimodal large language models: A survey},
  author={Wu, Junda and Zhang, Zhehao and Xia, Yu and Li, Xintong and Xia, Zhaoyang and Chang, Aaron and Yu, Tong and Kim, Sungchul and Rossi, Ryan A and Zhang, Ruiyi and others},
  journal={arXiv preprint arXiv:2409.15310},
  year={2024}
}

@inproceedings{
automatic_prompt_engineer,
title={Large Language Models are Human-Level Prompt Engineers},
author={Yongchao Zhou and Andrei Ioan Muresanu and Ziwen Han and Keiran Paster and Silviu Pitis and Harris Chan and Jimmy Ba},
booktitle={The Eleventh International Conference on Learning Representations },
year={2023},
url={https://openreview.net/forum?id=92gvk82DE-}
}

@misc{wu2025vtoolr1,
      title={VTool-R1: VLMs Learn to Think with Images via Reinforcement Learning on Multimodal Tool Use}, 
      author={Mingyuan Wu and Jingcheng Yang and Jize Jiang and Meitang Li and Kaizhuo Yan and Hanchao Yu and Minjia Zhang and Chengxiang Zhai and Klara Nahrstedt},
      year={2025},
      eprint={2505.19255},
      archivePrefix={arXiv},
      primaryClass={cs.LG},
      url={https://arxiv.org/abs/2505.19255}, 
}

@inproceedings{VisProg,
  title={Visual programming: Compositional visual reasoning without training},
  author={Gupta, Tanmay and Kembhavi, Aniruddha},
  booktitle={Proceedings of the IEEE/CVF conference on computer vision and pattern recognition},
  pages={14953--14962},
  year={2023}
}

@article{step-by-step,
  title={Large language models are zero-shot reasoners},
  author={Kojima, Takeshi and Gu, Shixiang Shane and Reid, Machel and Matsuo, Yutaka and Iwasawa, Yusuke},
  journal={Advances in neural information processing systems},
  volume={35},
  pages={22199--22213},
  year={2022}
}

@inproceedings{deepbreath,
  title={Large language models as optimizers},
  author={Yang, Chengrun and Wang, Xuezhi and Lu, Yifeng and Liu, Hanxiao and Le, Quoc V and Zhou, Denny and Chen, Xinyun},
  booktitle={The Twelfth International Conference on Learning Representations},
  year={2024}
}

@inproceedings{clip_visual_prompt_learning,
  title={Understanding and improving visual prompting: A label-mapping perspective},
  author={Chen, Aochuan and Yao, Yuguang and Chen, Pin-Yu and Zhang, Yihua and Liu, Sijia},
  booktitle={Proceedings of the IEEE/CVF Conference on Computer Vision and Pattern Recognition},
  pages={19133--19143},
  year={2023}
}

@article{clip_soft_visual_prompt,
  title={Learning visual prompts for guiding the attention of vision transformers},
  author={Rezaei, Razieh and Sabet, Masoud Jalili and Gu, Jindong and Rueckert, Daniel and Torr, Philip and Khakzar, Ashkan},
  journal={arXiv preprint arXiv:2406.03303},
  year={2024}
}

@misc{VLM_spatial_ambiguity,
      title={Do Vision-Language Models Represent Space and How? Evaluating Spatial Frame of Reference Under Ambiguities}, 
      author={Zheyuan Zhang and Fengyuan Hu and Jayjun Lee and Freda Shi and Parisa Kordjamshidi and Joyce Chai and Ziqiao Ma},
      booktitle={The Thirteenth International Conference on Learning Representations},
      year={2025},
      url={https://arxiv.org/abs/2410.17385}, 
}

@inproceedings{vlm_spatial_psychometrics,
    title = "Defining and Evaluating Visual Language Models' Basic Spatial Abilities: A Perspective from Psychometrics",
    author = "Xu, Wenrui  and
      Lyu, Dalin  and
      Wang, Weihang  and
      Feng, Jie  and
      Gao, Chen  and
      Li, Yong",
    editor = "Che, Wanxiang  and
      Nabende, Joyce  and
      Shutova, Ekaterina  and
      Pilehvar, Mohammad Taher",
    booktitle = "Proceedings of the 63rd Annual Meeting of the Association for Computational Linguistics (Volume 1: Long Papers)",
    month = jul,
    year = "2025",
    address = "Vienna, Austria",
    publisher = "Association for Computational Linguistics",
    url = "https://aclanthology.org/2025.acl-long.567/",
    doi = "10.18653/v1/2025.acl-long.567",
    pages = "11571--11590",
    ISBN = "979-8-89176-251-0",
}

@article{lost_in_the_middle,
  title={Lost in the middle: How language models use long contexts},
  author={Liu, Nelson F and Lin, Kevin and Hewitt, John and Paranjape, Ashwin and Bevilacqua, Michele and Petroni, Fabio and Liang, Percy},
  journal={Transactions of the association for computational linguistics},
  volume={12},
  pages={157--173},
  year={2024}
}

@article{grounding_dino,
  title={Grounding dino: Marrying dino with grounded pre-training for open-set object detection},
  author={Liu, Shilong and Zeng, Zhaoyang and Ren, Tianhe and Li, Feng and Zhang, Hao and Yang, Jie and Jiang, Qing and Li, Chunyuan and Yang, Jianwei and Su, Hang and Zhu, Jun and Zhang, Lei},
  journal={ECCV},
  year={2024}
}

@article{semantic_sam,
  title={Semantic-SAM: Segment and Recognize Anything at Any Granularity},
  author={Li, Feng and Zhang, Hao and Sun, Peize and Zou, Xueyan and Liu, Shilong and Yang, Jianwei and Li, Chunyuan and Zhang, Lei and Gao, Jianfeng},
  journal={ECCV},
  year={2024}
}

@inproceedings{depthanything,
  title={Depth Anything: Unleashing the Power of Large-Scale Unlabeled Data},
  author={Yang, Lihe and Kang, Bingyi and Huang, Zilong and Xu, Xiaogang and Feng, Jiashi and Zhao, Hengshuang},
  booktitle={CVPR},
  year={2024}
}

@InProceedings{uct,
author="Kocsis, Levente
and Szepesv{\'a}ri, Csaba",
editor="F{\"u}rnkranz, Johannes
and Scheffer, Tobias
and Spiliopoulou, Myra",
title="Bandit Based Monte-Carlo Planning",
booktitle="Machine Learning: ECML 2006",
year="2006",
publisher="Springer Berlin Heidelberg",
address="Berlin, Heidelberg",
pages="282--293",
isbn="978-3-540-46056-5"
}

@article{ucb,
  title={Finite-time analysis of the multiarmed bandit problem},
  author={Auer, Peter and Cesa-Bianchi, Nicolo and Fischer, Paul},
  journal={Machine learning},
  volume={47},
  number={2},
  pages={235--256},
  year={2002},
  publisher={Springer}
}

@inproceedings{blackbox_vp,
    title = "Black-Box Visual Prompt Engineering for Mitigating Object Hallucination in Large Vision Language Models",
    author = "Woo, Sangmin  and
      Zhou, Kang  and
      Zhou, Yun  and
      Wang, Shuai  and
      Guan, Sheng  and
      Ding, Haibo  and
      Cheong, Lin Lee",
    editor = "Chiruzzo, Luis  and
      Ritter, Alan  and
      Wang, Lu",
    booktitle = "Proceedings of the 2025 Conference of the Nations of the Americas Chapter of the Association for Computational Linguistics: Human Language Technologies (Volume 2: Short Papers)",
    month = apr,
    year = "2025",
    address = "Albuquerque, New Mexico",
    publisher = "Association for Computational Linguistics",
    url = "https://aclanthology.org/2025.naacl-short.45/",
    doi = "10.18653/v1/2025.naacl-short.45",
    pages = "529--538",
    ISBN = "979-8-89176-190-2"
}

@article{vtsv,
          title={Multi-Step Visual Reasoning with Visual Tokens Scaling and Verification},
          author={Bai, Tianyi and Hu, Zengjie and Sun, Fupeng and Qiu, Jiantao and Jiang, Yizhen and He, Guangxin and Zeng, Bohan and He, Conghui and Yuan, Binhang and Zhang, Wentao},
          journal={NeurIPS},
          year={2025},
          url={https://arxiv.org/abs/2506.07235},
        }

\newpage
\appendix

\setcounter{table}{0}  
\setcounter{figure}{0}
\renewcommand{\thetable}{A\arabic{table}}
\renewcommand{\thefigure}{A\arabic{figure}}

\section*{Appendix}

\section{Prompts for each step}
\label{appendix:prompt_details}

We provide the prompts used for SEVEX.

\begin{tcolorbox}[title={\textbf{\small Implementation prompt}}, boxrule=2pt, arc=0mm, breakable]\begin{minted}[fontsize=\scriptsize, breaklines, breakanywhere, frame=lines, framesep=2mm, tabsize=4, style=vs, autogobble]{python}
PROMPT = """You are a machine learning engineer. Your task is to analyze the problem and previous attempts, then propose a new improvement idea. 
Another agent will update the code and prompt to implement the idea later.

Problem Description:
{problem_description}

Previous ideas:
{parent_idea}

{sibling_ideas}

Implications from Previous Experiments:
{parent_implications}


Based on the problem description and the implications from the parent experiment, propose a new idea or strategy to solve the given problem. Consider:
- What worked well based on the parent experiment implications?
- What didn't work well according to the implications?
- What new approaches could be tried based on these implications?
- How can we leverage the available functions more effectively?

IMPORTANT: Your idea must be clearly different from the sibling ideas listed above. Generate a promising idea that builds on the parent experiment's implications while exploring a distinct direction from what siblings have already tried.

Provide a clear, concise idea description (2-4 sentences) that explains the strategy you want to try next. Make sure the idea is simple and concise to make gradual improvement.
The idea should include:
- Plan about which changes to make
- What are the expected results and changes in the result image
- How to evaluate if the idea is successful or not
Which should be excluded in the response:
- Previous failed trials
- Verbose explanation
- Explicit code snippet


Available Functions (signatures + summaries):
{functions_reference}
"""
\end{minted}
\end{tcolorbox}

\begin{tcolorbox}[title={\textbf{\small Sample-wise analysis prompt}}, boxrule=2pt, arc=0mm, breakable]\begin{minted}[fontsize=\scriptsize, breaklines, breakanywhere, frame=lines, framesep=2mm, tabsize=4, style=vs, autogobble]{python}
PROMPT = """You are analyzing whether an idea was implemented and used correctly for this sample.

Context:
- Idea (intent of the pseudo-code): {idea}
- Prediction (model output for this sample): {prediction}
- Ground truth: {ground_truth}
- Error message (if any): {error}

Image(s) attached in order:
1. Input image for this sample (required)
2. Last image sent to the VLM for answering (optional; if only one image is attached, this was not provided)

Tasks:
(a) Assess whether the idea appears to be implemented and used correctly for this sample given the image(s) and the prediction vs ground truth.
(b) If there was an error or the prediction is wrong, suggest likely causes (e.g., wrong region used, VLM misinterpretation, code bug, missing preprocessing).

Respond in the following format (plain text, no markdown bullets):
IMPLICATIONS: <1-2 sentences on whether the idea was applied correctly and how it influenced the result>
CAUSES: <if error or wrong prediction: 1-2 short bullet-like causes; otherwise write "None">
"""
\end{minted}
\end{tcolorbox}

\begin{tcolorbox}[title={\textbf{\small Insights generation prompt}}, boxrule=2pt, arc=0mm, breakable]\begin{minted}[fontsize=\scriptsize, breaklines, breakanywhere, frame=lines, framesep=2mm, tabsize=4, style=vs, autogobble]{python}
PROMPT = """You are an expert in analyzing experimental results. Your task is to summarize the execution results and extract key implications.

Execution Results:
- Success: {success}
- Reward: {reward_str}
- Error: {error}

Idea for This Iteration:
{idea}

Per-Image Analysis:
{image_comparisons}

Provide two outputs:
1. A concise summary (2-3 sentences) of what happened during execution and the results
   - Reference how the stated idea influenced the outcome (success/failure, partial progress)
   - Reference the per-image comparisons when noting visual problems or successes
2. Key implications (2-4 bullet points) about what worked, what didn't, and what could be improved
   - Highlight which images had the most severe differences

Format your response as:
SUMMARY:
[your summary here]

IMPLICATIONS:
- [implication 1]
- [implication 2]
- [implication 3]

"""
\end{minted}
\end{tcolorbox}

\begin{tcolorbox}[title={\textbf{\small Insights back propagation prompt}}, boxrule=2pt, arc=0mm, breakable]\begin{minted}[fontsize=\scriptsize, breaklines, breakanywhere, frame=lines, framesep=2mm, tabsize=4, style=vs, autogobble]{python}
PROMPT = """You are an expert in synthesizing experimental insights. Your task is to revise and consolidate implications by combining the current node's implications with insights from its children's experiments.

Current Node's Implications:
{current_implication}

Children's Implications:
{children_implications}

Your task:
1. Synthesize the key insights from both the current node and its children
2. Identify patterns, common themes, and important learnings
3. Generate a revised, consolidated list of implications
4. Focus on actionable insights that can guide future exploration

Requirements:
- Output must be a bulletized list (using "- " prefix)
- Maximum 5 bullet points
- Each bullet should be concise but informative
- Prioritize the most important and actionable insights
- Remove redundancy and merge similar points

Format your response as a bulletized list:
- [revised implication 1]
- [revised implication 2]
- [revised implication 3]
- [revised implication 4]
- [revised implication 5]

"""
\end{minted}
\end{tcolorbox}

\begin{tcolorbox}[title={\textbf{\small Idea generation prompt}}, boxrule=2pt, arc=0mm, breakable]\begin{minted}[fontsize=\scriptsize, breaklines, breakanywhere, frame=lines, framesep=2mm, tabsize=4, style=vs, autogobble]{python}
PROMPT = """You are a machine learning engineer. Your task is to analyze the problem and previous attempts, then propose a new improvement idea. 
Another agent will update the code and prompt to implement the idea later.

Problem Description:
{problem_description}

Previous ideas:
{parent_idea}

{sibling_ideas}

Implications from Previous Experiments:
{parent_implications}


Based on the problem description and the implications from the parent experiment, propose a new idea or strategy to solve the given problem. Consider:
- What worked well based on the parent experiment implications?
- What didn't work well according to the implications?
- What new approaches could be tried based on these implications?
- How can we leverage the available functions more effectively?

IMPORTANT: Your idea must be clearly different from the sibling ideas listed above. Generate a promising idea that builds on the parent experiment's implications while exploring a distinct direction from what siblings have already tried.

Provide a clear, concise idea description (2-4 sentences) that explains the strategy you want to try next. Make sure the idea is simple and concise to make gradual improvement.
The idea should include:
- Plan about which changes to make
- What are the expected results and changes in the result image
- How to evaluate if the idea is successful or not
Which should be excluded in the response:
- Previous failed trials
- Verbose explanation
- Explicit code snippet


Available Functions (signatures + summaries):
{functions_reference}
"""
\end{minted}
\end{tcolorbox}

\begin{tcolorbox}[title={\textbf{\small Idea self-evaluation prompt}}, boxrule=2pt, arc=0mm, breakable]\begin{minted}[fontsize=\scriptsize, breaklines, breakanywhere, frame=lines, framesep=2mm, tabsize=4, style=vs, autogobble]{python}
PROMPT = """You are a computer science engineer. Your task is to evaluate an improvement idea on three dimensions: feasibility, expectation, and novelty.

Idea to Evaluate:
{idea}

Sibling ideas:
{sibling_ideas}

Available Functions (signatures + summaries):
{functions_reference}

Evaluate the idea on the following dimensions:

1. **Feasibility** (1-5): Can this idea be implemented using only the available functions listed above?
   - 5: Fully implementable with available functions, no additional capabilities needed
   - 4: Mostly implementable, may require creative use of available functions
   - 3: Partially implementable, some aspects may be challenging with available functions
   - 2: Difficult to implement, requires additional functions and imports
   - 1: Not implementable with available functions

2. **Expectation** (1-5): How confident are you that this idea will make significant improvement to the performance?
   - 5: Very high confidence, idea directly addresses key performance bottlenecks
   - 4: High confidence, idea addresses important aspects of the problem
   - 3: Moderate confidence, idea may provide some improvement
   - 2: Low confidence, idea is unlikely to provide significant improvement
   - 1: Very low confidence, idea is unlikely to help

3. **Novelty** (1-5): How different is this idea from the sibling ideas listed above?
   - 5: Completely different approach, explores new direction
   - 4: Significantly different, with some unique aspects
   - 3: Moderately different, some overlap with siblings
   - 2: Similar to sibling ideas, minor variations
   - 1: Very similar to sibling ideas, minimal differentiation

Return ONLY a JSON object with scores (1-5) for each dimension, without any explanations:
{{
  "feasibility": <1-5>,
  "expectation": <1-5>,
  "novelty": <1-5>
}}

"""
\end{minted}
\end{tcolorbox}

\section{Details of datasets}
\label{appendix:dataset_details}

\subsection{BlindTest: Vision Language Models are Blind}
The BlindTest benchmark focuses on evaluating the fundamental geometric perception of Vision-Language Models (VLMs). It utilizes simple 2D vector graphics to ensure that models cannot rely on language priors or contextual cues, forcing them to perform pure visual reasoning.

\begin{itemize}
\item \textbf{Counting Line Intersections:} This task requires the model to determine the exact number of intersection points (typically 0, 1, or 2) between two piece-wise linear function segments. It tests the model's precision in pixel-level coordinate alignment.
\item \textbf{The Circled Letter:} Given a string of characters where a single letter is enclosed in a red circle, the model is asked to identify the specific character. This evaluates the model's ability to associate a spatial container (the circle) with a specific semantic token (the letter).
\item \textbf{Following Single-colored Paths (Subway Map):} Inspired by subway maps, this task involves tracing a specific colored path from a starting point to an endpoint within a complex graph. It assesses the connectivity and path-tracing capabilities of LVLMs.
\item \textbf{Counting Overlapping Shapes:} The model is presented with two or more circles and must determine their topological relationship, such as whether they overlap, touch, or maintain a subtle gap. This tests the understanding of spatial boundaries and containment.
\end{itemize}

\subsection{BLINK: Multimodal Large Language Models Can See but Not Perceive}
The BLINK benchmark consists of 14 diverse visual tasks that are trivial for humans (solvable "in a blink") but challenging for current VLMs. These tasks are designed to be "un-captionable," meaning they cannot be solved using text-only metadata.
We use five tasks which are used in SketchPad.

\begin{itemize}
\item \textbf{Jigsaw:} This task involves reassembling image tiles into their original structure. It evaluates the model's grasp of global structural consistency and local texture matching.
\item \textbf{Relative Depth:} Given two points (A and B) in a single monocular image, the model is asked to judge which point is physically further from the camera. This measures the model's ability to reconstruct 3D spatial cues from 2D projections.
\item \textbf{Spatial Relation:} This task tests the understanding of 3D spatial arrangements (e.g., "left-of", "behind", "on-top-of") between multiple objects in a scene, going beyond simple 2D bounding box detection.
\item \textbf{Semantic Correspondence:} The model is asked to identify semantically similar yet visually distinct parts across different instances of a category (e.g., the "beak" of two different bird species). This requires high-level semantic abstraction of visual features.
\item \textbf{Visual Correspondence:} Unlike semantic matching, this task requires finding the exact same physical point across two different viewpoints, lighting conditions, and time of the same scene. It assesses the model's robustness to geometric transformations and viewpoint changes.
\end{itemize}

\subsection{Indexes for development set}

\begin{tcolorbox}[title={\textbf{\small Indices for development set}}, boxrule=2pt, arc=0mm, breakable]\begin{minted}[fontsize=\scriptsize, breaklines, breakanywhere, frame=lines, framesep=2mm, tabsize=4, style=vs, autogobble]{python}
{
"intersection": [1153,1290,1354,1623,1660,1740,1761,1989,2016,2274,2300,2459,2694,2731,2751,2823,2958,2989,2990,3220,3373,3608,3620,3643,3816,3901,4023,4146,4300,4517],
"circledletter": [7415,7431,7438,7441,7453,7470,7473,7474,7495,7517,7530,7533,7559,7602,7698,7700,7704,7734,7788,7789,7807,7808,7847,7851,7854,7898,7902,7909,7956,7992],
"subway": [21,73,75,91,98,128,168,194,212,233,241,253,294,307,326,342,349,372,383,408,557,574,582,583,622,629,650,699,702,705],
"overlappingshapes": [6442,6466,6501,6516,6518,6530,6533,6546,6552,6554,6587,6594,6602,6619,6691,6705,6744,6758,6795,6799,6801,6808,6829,6841,6847,6849,6856,6861,6882,6895],
"jigsaw": [6,13,15,20,26,29,32,39,47,53,55,68,70,82,83,90,92,93,100,101,108,111,122,126,127,128,129,131,139,145],
"depth": [3,9,13,15,17,22,23,24,26,30,33,38,44,48,49,53,58,60,66,67,72,82,87,92,93,107,110,115,118,122],
"spatial": [1,5,7,10,11,23,25,29,32,40,42,43,46,51,63,68,72,82,84,86,93,99,103,104,109,110,113,124,133,142],
"semcorr": [11,17,22,27,31,34,35,40,42,44,48,56,57,66,70,79,80,84,88,90,100,105,113,125,128,131,136,137,138,139],
"viscorr": [6,20,22,32,35,42,45,46,51,55,59,60,68,72,74,87,88,91,102,103,109,111,114,121,136,142,147,148,155,158]
}
\end{minted}
\end{tcolorbox}




\section{Additional analyses}

\subsection{Restricted-toolbox ablation}
\label{appendix:restricted_toolbox}

To verify that the gains of SEVEX stem from the semantic search over visual strategies rather than from access to powerful external pretrained tools, we restrict the toolbox to only the basic and drawing primitives in Table~\ref{tab:visual_tools} (i.e., \texttt{get\_image\_size}, \texttt{convert\_image\_grayscale}, \texttt{crop}, \texttt{overlay\_images}, \texttt{draw\_line}, \texttt{draw\_box}, \texttt{draw\_filled\_box}), excluding the external-model tools (\texttt{detect\_objects}, \texttt{sliding\_window\_detection}, \texttt{segment\_and\_mark}, \texttt{estimate\_depth}).
As shown in Table~\ref{tab:restricted_toolbox}, SEVEX with the restricted toolbox still substantially improves over the Naive baseline (average 65.6\% $\rightarrow$ 73.3\%) and remains comparable to the full-tool SEVEX (72.4\%).
This indicates that complex external tools are not the main source of the BlindTest gains.
External tools nonetheless help on some tasks; for example, on OverlappingShapes the restricted toolbox (45.6\%) trails the full-tool variant (52.4\%), since detection and segmentation expose relations that the basic primitives cannot.

\begin{table}[t]
\caption{
    \textbf{Restricted-toolbox ablation on BlindTest (Gemini-2.5-flash).}
    Test accuracy (\%) when SEVEX is restricted to basic and drawing tools only, compared to the Naive baseline and the full-tool SEVEX from Table~\ref{tab:main_table}.
}
\label{tab:restricted_toolbox}
\centering
\begin{tabular}{lccc}
\toprule
\textbf{Task} & Naive & Basic-tool SEVEX & Full-tool SEVEX \\
\midrule
LineIntersections & 73.0 & \textbf{91.1} & 90.5 \\
CircledLetter & 78.8 & \textbf{90.9} & 83.7 \\
SubwayMap & 60.0 & \textbf{65.4} & 62.8 \\
OverlappingShapes & 50.4 & 45.6 & \textbf{52.4} \\
\midrule
\textbf{Average} & 65.6 & \textbf{73.3} & 72.4 \\
\bottomrule
\end{tabular}
\end{table}

\subsection{Sensitivity to the development split}
\label{appendix:dev_split}

A small 30-image development set raises the concern that the discovered prompt might overfit the specific development samples.
To examine this, we construct two additional random development splits of 30 images each for the CircledLetter task, and reserve a test set that excludes the union of images used by these splits.
We run SEVEX five times per split and report the mean and within-split standard deviation in Table~\ref{tab:dev_split}.
Across splits, the held-out test accuracy stays in the mid-to-high 80s, well above the Naive baseline of 78.8\%, supporting that the discovered strategy is not merely development-set memorization.
This is consistent with the smaller development--test gap of SEVEX relative to SketchPad$+$APE reported in the main paper.
We limit this study to CircledLetter and do not claim split stability for all tasks.

\begin{table}[t]
\caption{
    \textbf{Random development-split sensitivity on CircledLetter (Gemini-2.5-flash).}
    Accuracy (\%) reported as mean $\pm$ standard deviation over five runs per split. The Naive baseline accuracy is 78.8\%.
}
\label{tab:dev_split}
\centering
\begin{tabular}{lcc}
\toprule
\textbf{Set} & Split 1 & Split 2 \\
\midrule
CircledLetter dev & $90.20 \pm 3.00$ & $92.64 \pm 0.88$ \\
CircledLetter test & $88.00 \pm 4.30$ & $85.98 \pm 3.86$ \\
\bottomrule
\end{tabular}
\end{table}

\subsection{Step-by-step search analysis}
\label{appendix:step-by-step}

\begin{figure}[t]
    \centering    
    \includegraphics[width=0.8\linewidth]{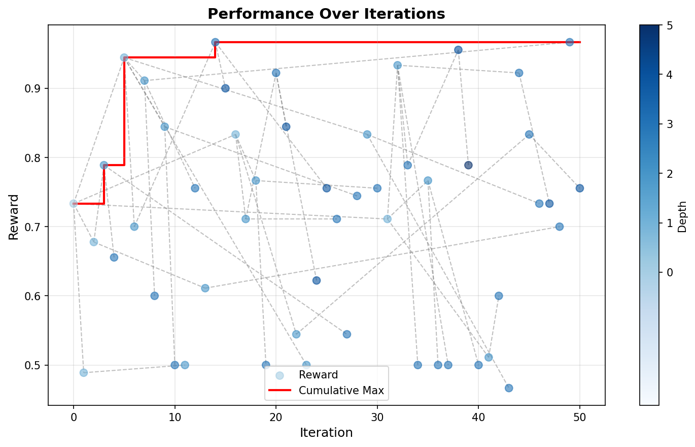}
    \caption{
        \textbf{Example intermediate rewards over search iterations.}
        Each node denotes a visual prompt trial, and dotted lines indicate parent-child relations.
        SEVEX explores diverse candidates while revisiting promising nodes for further refinement.
    }
    \label{fig:intermediate_rewards}
\end{figure}

Figure~\ref{fig:intermediate_rewards} provides a representative trace of the SEVEX search process, illustrating how the method balances broad exploration with repeated refinement of promising directions.
Each node corresponds to a visual prompt trial, and dotted edges indicate parent-child relations formed when SEVEX expands a previous idea.
A node does not need to obtain a high reward in its first trial to remain useful: if its analysis suggests a promising direction, SEVEX can still generate child nodes that revise or specialize the idea.
Thus, the search is not a simple greedy selection over immediate rewards; the exploration hyperparameters encourage diversity while allowing high-potential branches to be revisited.

\section{Discussions}
    \label{sec:discussion}

    \subsection{Will the visual perception failures of LVLMs disappear in future models?}

        The inherent opacity of large-scale models makes it fundamentally difficult to fully understand or predict their behavioral trajectories. 
        While future models may improve, the image perception issues we have addressed represent a broader challenge in model reliability. 
        We have demonstrated that the engineering of visual prompts can be effectively automated through an agentic system that leverages high-level semantic exploration. 
        This is evidenced by SEVEX’s ability to utilize visual tools in creative ways that extend beyond their intended purposes. 
        For example, in the Jigsaw task provided in Figure~\ref{fig:qualitative_prompt}, the agent utilized a depth estimation model to judge the naturalness of an image by identifying sharp, unnatural discontinuities, which is a counter-intuitive solution that a human-designed heuristic might overlook. 
        Such emergent tool-use behavior highlights the potential of agent-driven exploration for solving increasingly complex reasoning tasks.
    


    \subsection{How can human-AI collaboration further enhance the result?}
        Integrating human expertise into the semantic backpropagation loop presents a promising direction for future research. By allowing researchers to provide high-level directional guidance, the exploration tree can be refined more efficiently, pruning ineffective branches based on domain knowledge. This synergy between automated empirical discovery and human intuition establishes a new paradigm for enhancing both the transparency and reliability of LVLMs.

    
    \subsection{Can SEVEX surpass human engineers?}
        The visual prompt of Jigsaw, shown in Figure~\ref{fig:qualitative_prompt}, is an example of counter-intuitive but effective visual prompts. 
        It shows potential of finding solutions which is not easily imagined by human engineer. 
        Not only reducing the human burden in method finding, but also it could provide solutions beyond the human expectation.

        
        


\end{document}